\documentclass[runningheads]{llncs}

\usepackage{eccv}

\usepackage{eccvabbrv}

\usepackage{graphicx}
\usepackage{booktabs}

\usepackage[accsupp]{axessibility}  
\usepackage{url}

\usepackage[pagebackref,breaklinks,colorlinks,citecolor=eccvblue]{hyperref}
\usepackage{hyperref}

\begin{document}

\title{Forgedit: Text-guided Image Editing  via Learning and Forgetting} 

\titlerunning{Forgedit: Text-guided Image Editing  via Learning and Forgetting}

\author{Shiwen Zhang\inst{1} \and
Shuai Xiao\inst{2}and
Weilin Huang\inst{2}}

\authorrunning{Shiwen Zhang et al.}

\institute{Bytedance \\
\email{shiwen.zhang@bytedance.com}\and
Alibaba\\
\email{shuai.xsh@alibaba-inc.com}\\\email{whuang@robots.ox.ac.uk}
}

\maketitle

\begin{abstract}
 Text-guided image editing on real or synthetic images, given only the original image itself and the target text prompt as inputs, is a very general and challenging task. It requires an editing model to estimate by itself which part of the image should be edited, and then perform  either rigid or non-rigid editing while preserving the characteristics of original image.   In this paper, we design a novel text-guided image editing method, named as Forgedit. First, we propose a vision-language joint optimization framework capable of reconstructing the original image in 30 seconds, much faster than previous SOTA and much less overfitting.  Then we propose a novel vector projection mechanism in text embedding space of Diffusion Models, which is capable to control the identity similarity and editing strength seperately. Finally, we discovered a general property of UNet in Diffusion Models, i.e., Unet encoder learns space and structure, Unet decoder learns appearance and identity. With such a property, we design forgetting mechanisms to successfully tackle the fatal and inevitable overfitting issues when fine-tuning Diffusion Models on one image, thus significantly boosting the editing capability of Diffusion Models. Our method, Forgedit, built on Stable Diffusion, achieves new state-of-the-art results on the challenging text-guided image editing benchmark: TEdBench,  surpassing the previous SOTA methods such as Imagic with Imagen, in terms of both CLIP score and LPIPS score.

\end{abstract}

\section{introduction}
Image Editing \cite{DBLP:conf/siggraph/OhCDD01} is a fundamental problem in computer vision. 
To perform image editing, a guidance condition is often provided to the model to indicate the editing intention.
Text condition is the most simple and general form of such editing guidance, in which case the editing task is called text-guided image editing. Such a text describing the content of the desired edited image is usually called  target prompt. 
In this paper, we aim to tackle the task of text-guided image editing, with only an original image and a target prompt provided - which are the minimum requirements of inputs for this task.  Text-guided image editing is a general task, including  both rigid and non-rigid editing.

 The approaches of text-guided image editing are generally categorized into optimization-based methods and non-optimization ones according to whether  fine-tuning process is performed for reconstruction. 
Recent non-optimization editing methods \cite{brooks2022instructpix2pix,Tumanyan_2023_CVPR,DBLP:conf/iclr/CouaironVSC23,zhang2023adding,Meng2021SDEditGI,DBLP:conf/cvpr/AvrahamiLF22,DBLP:conf/iccv/CaoWQSQZ23,wei2023elite} are very efficient. Yet they
%
either struggle on  preserving the precise characteristics of original image during complex editing, or suffer from being incapable of performing sophisticated and accurate non-rigid edits.
It is undeniable that  fine-tuning a diffusion model with the original image is still critical and  necessary for precise identity preservation and accurate semantic understanding.   However, previous optimization-based methods \cite{DBLP:conf/cvpr/KawarZLTCDMI23,DBLP:conf/cvpr/RuizLJPRA23,zhang2022sine} suffer from long fine-tuning time, severe overfitting issues or incapablity of performing precise non-rigid editing. 
\begin{figure*}[!h]
  \centering
  
   \includegraphics[width=0.8\linewidth]{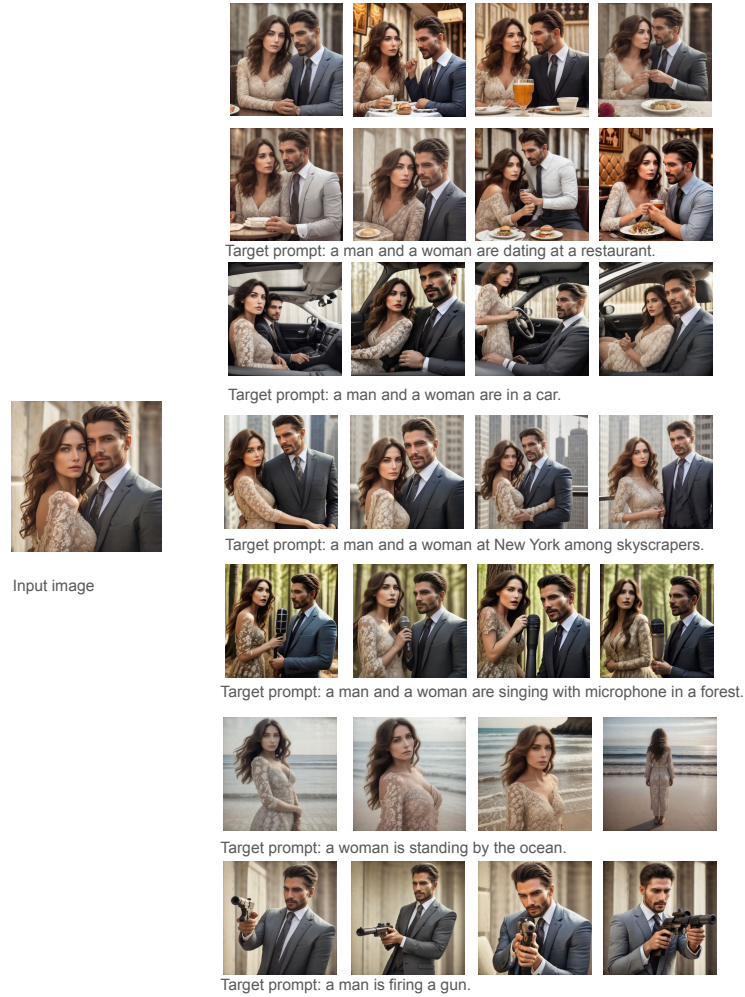}
\vspace{-10pt}
   \caption{ Forgedit could be used for consistent and controllable keyframe generation for visual storytelling and movie generation,  given one input image and target prompts. We list several samples with different random seeds for each target prompt. We demonstrate Forgedit is capable of controling multiple characters performing various actions at different scenes. Forgedit could also control each different character seperately. Forgetting strategy on UNet's encoder  with vector subtraction leads to high flexibility and success rate  to change the spatial structures and actions,  preserving appearance and identity by reserving  UNet's decoder.
   }
   \label{movie}
\end{figure*}
In this paper, we are going to tackle the aforementioned issues of optimization-based editing methods.  We name our text-guided image editing method as \textit{Forgedit} (similar to \textit{forget it}), which consists of two stages: fine-tuning and editing.

For fine-tuning stage, with a generated source prompt from BLIP \cite{Li2022BLIPBL} to describe the original image, we design a vision and language joint optimization  framework,  which could be regarded as a variant of Imagic\cite{DBLP:conf/cvpr/KawarZLTCDMI23} by combining the first stage and the second stage of Imagic into one  and using BLIP generated caption as source prompt instead of using target prompt as source prompt like what Imagic does. Such simple modifications are the keys to  much faster convergence speed and less overfitting than Imagic.  With our joint learning of image and source text embedding, the finetuning stage using one image with our Forgedit+Stable Diffusion 1.4 \cite{Rombach2021HighResolutionIS} takes 30 seconds on an A100 GPU, compared with 7 minutes with Imagic +Stable Diffusion \cite{DBLP:conf/cvpr/KawarZLTCDMI23} reported by Imagic paper. This leads to  14x  speed up.   BLIP generated source prompt also eases overfitting, which we will demonstrate in the ablation study.

For editing stage, we propose two novel methods, vector projection and forgetting strategy with a finding of a general UNet property.  For the first time in the literature of text-guided image editing with Diffusion Models, we propose a novel vector projection mechanism in text embedding space of Diffusion Models, which is capable to seperately control the identity and editing strength by decomposing the target text embedding into identity embedding and editing embedding. We explore the properties of vector projection and compare it with previous vector subtraction method utilized in Imagic to demonstrate its superiority on identity preservation. Finally, we discovered a general property of UNet structure in Diffusion Models, i.e., Unet encoder learns space and structure, Unet decoder learns appearance and identity. With such a property, we could easily tackle the  fatal overfitting issues of optimization-based image editing methods in a very effective and efficient manner during sampling process instead of fine-tuning process,  by designing a forgetting mechanism according to such a UNet property.

To sum up, our main contributions are:\\
1. We present Forgedit, an efficient vision-language joint optimization  framework, capable of performing both rigid and non-rigid text-guided image editing, while effectively tackling multiple general challenges on this task.\\
2. We introduce a novel vector projection mechanism in text embedding space of Diffusion Models, which provides a more controllable alternative for computing the combination of the source text embedding and the target text embedding. This improves Forgedit's capability for preserving more consistent characteristics of original image than existing methods.\\
3. We design a novel forgetting strategy based on our observation on the UNet architecture of diffusion models, i.e., Unet encoder learns space and structure, Unet decoder learns appearance and identity. This allows us to effectively tackle the critical overfitting issue on optimization based image editing methods,  thus significantly boosting the editing capability of diffusion models. 

Our Forgedit can achieve new state-of-the-art results on the challenging benchmark TEdBench  \cite{DBLP:conf/cvpr/KawarZLTCDMI23} (even by using an outdated Stable Diffusion 1.4), surpassing previous SOTA Imagic built on Imagen in terms of both CLIP score \cite{DBLP:conf/emnlp/HesselHFBC21} and LPIPS score \cite{Zhang2018TheUE}. Our Forgedit is highly flexible, and is readily applicable to other fine-tuning based text-guided image editing methods, with significant performance improvements. We will show more applications for such extensions in the appendix.
\vspace{-10pt}
\section{related work}
{\bf Text to Image Diffusion Models} Diffusion Models have dominated text to image generation. DDPM\cite{DBLP:conf/nips/HoJA20} improves Diffusion process proposed by \cite{DBLP:conf/icml/Sohl-DicksteinW15} on generating images. DDIM \cite{DBLP:conf/iclr/SongME21} accelerates the sampling procedure of Diffusion Models by making reverse process deterministic and using sub-sequence of time-steps. Dalle 2 \cite{ramesh2022hierarchical} trains a diffusion prior to convert a text caption to CLIP \cite{Radford2021LearningTV} image embedding and then employs a Diffusion Decoder to transfer the generated CLIP image embedding to  an image.  Imagen \cite{DBLP:conf/nips/SahariaCSLWDGLA22} is a Cascaded Diffusion Model \cite{Ho2021CascadedDM}, whose UNet is composed of three Diffusion Models generating images with increasing resolutions, employing the powerful T5 text encoder \cite{DBLP:journals/jmlr/RaffelSRLNMZLL20} for complex semantic understanding and generating sophisticated scenarios. Stable Diffusion \cite{Rombach2021HighResolutionIS} utilizes Variational AutoEncoders \cite{DBLP:journals/corr/KingmaW13} to compress the training image to a compact latent space so that the UNets could be trained with low resolution latents in order to save computational resources.

{\bf   Image Editing with Diffusion Models} Empowered by recent progress in text-to-image Diffusion Models, image editing methods have witnessed remarkable improvements.  ControlNets \cite{zhang2023adding} are trained on extra datasets to learn generating images with different conditions. However, these conditions only reflect partial attributes of the original image thus ControlNets are incapable of preserving the identity of the object being edited and also struggle to conduct non-rigid edits.  ELITE\cite{wei2023elite} utilize extra image encoder to transfer the appearance of objects yet fail to keep precise identity. Inpainting Models based on Diffusion Models, for example, Blended Diffusion\cite{DBLP:conf/cvpr/AvrahamiLF22}, require masks to indicate the editing region. They cannot preserve the identity of the object being edited and  cannot conduct non-rigid editing due to the restricts of the region of masks,  for example, making a bird perching on the branch spread its wings.  SDEdit \cite{Meng2021SDEditGI} utilizes DDIM Inversion to add noises to the original image and then denoises the image with target prompt. DiffEdit \cite{DBLP:conf/iclr/CouaironVSC23} obtains the target object mask with Diffusion Model itself by a user provided source prompt and conduct SDEdit in the mask region. PnP Diffusion \cite{Tumanyan_2023_CVPR} injects intermediate features of original image to the generation of target prompt. Instruct pix2pix \cite{brooks2022instructpix2pix} pretrains the Diffusion Models on external datasets with triplets of original image, edited image and target prompt. All these non-optimization methods  suffer from  the fact that they are either incapable of preserving the precise characteristics or unable to conduct complex non-rigid editing. Drag Diffusion \cite{DBLP:journals/corr/abs-2307-02421}, which extends DragGAN \cite{DBLP:conf/siggraph/PanTLLMT23} to Diffusion Models, is capable of preserving the identity of the original image and performs non-rigid edits. However, Drag Diffusion and DragGAN are only capable of performing space-related editing, which is just a portion of general image editing tasks. Instead, our Forgedit is a general text-guided image editing framework to conduct various kinds of image editing operations, including spatial transformations. Prompt to Prompt \cite{DBLP:conf/iclr/HertzMTAPC23} requires that the source prompt and target prompt must be provided in a precisely word-by-word matching form so that the algorithm could accurately find the editing target, which is  too ideal thus impossible in our setting. Imagic \cite{DBLP:conf/cvpr/KawarZLTCDMI23} is a three-stage optimization based editing method, which is the current state-of-the-art text-guided image editing algorithm, which could be regarded as a combination of a variant of textual inversion \cite{DBLP:conf/iclr/GalAAPBCC23}in the first stage and DreamBooth \cite{DBLP:conf/cvpr/RuizLJPRA23}in the second stage. However, the fine-tuning stages of Imagic are very slow and suffer from overfitting.
\label{related work}
\vspace{-10pt}
\section{Forgedit}

\subsection{Preliminaries}
 Diffusion models \cite{DBLP:conf/nips/HoJA20,DBLP:conf/icml/Sohl-DicksteinW15} consist of a forward process and a reverse process. The forward process starts from the given image  $x_0$, and  then progressively add  Gaussian Noise $\epsilon_t \sim \mathcal{N}(0, 1)$ in each timestep $t$ to get $x_t$. In such a diffusion process, $x_t$ can be directly calculated at each timestep $t \in \{0,...,T\}$,
 \begin{equation}
 x_t=\sqrt{\alpha _t}x_0+\sqrt{1-\alpha _t}\epsilon_t
\label{eq one channel normalized}
\end{equation}
with $\alpha _t$ being diffusion schedule parameters with $0=\alpha_T<\alpha_{T-1}...<\alpha_1<\alpha_0=1$ .

In the reverse process, given $x_t$ and text embedding $e$, the time-conditional UNets $\epsilon_\theta(x_t,t,e)$ of diffusion models predict random noise $\epsilon_t$ added to $x_{t-1}$. With DDIM \cite{DBLP:conf/iclr/SongME21}, the reverse process can be computed as,
 \begin{equation}
 x_{t-1}=\frac{\sqrt{\alpha_{t-1}}}{\sqrt{\alpha_t}}(x_t-\sqrt{1-\alpha_t} \epsilon_\theta(x_t,t,e))+\sqrt{1-\alpha_{t-1}}\epsilon_\theta(x_t,t,e)
\label{eq one channel normalized}
\end{equation}

With Latent Diffusion Models \cite{Rombach2021HighResolutionIS}, the original image $x_0$ is replaced by a latent representation $z_0$ obtained from  a VAE \cite{DBLP:journals/corr/KingmaW13} Encoder $\varepsilon(x_0)$. The overall training loss is computed as, 

\begin{equation}
L = \mathbb{E}_{z_t,\epsilon_t,t,e} ||\epsilon_t -\epsilon_\theta(z_t,t,e)||_2^2
\label{eq loss}
\end{equation}

 \begin{figure*}[t]
  \centering
   \includegraphics[width=0.8\linewidth]{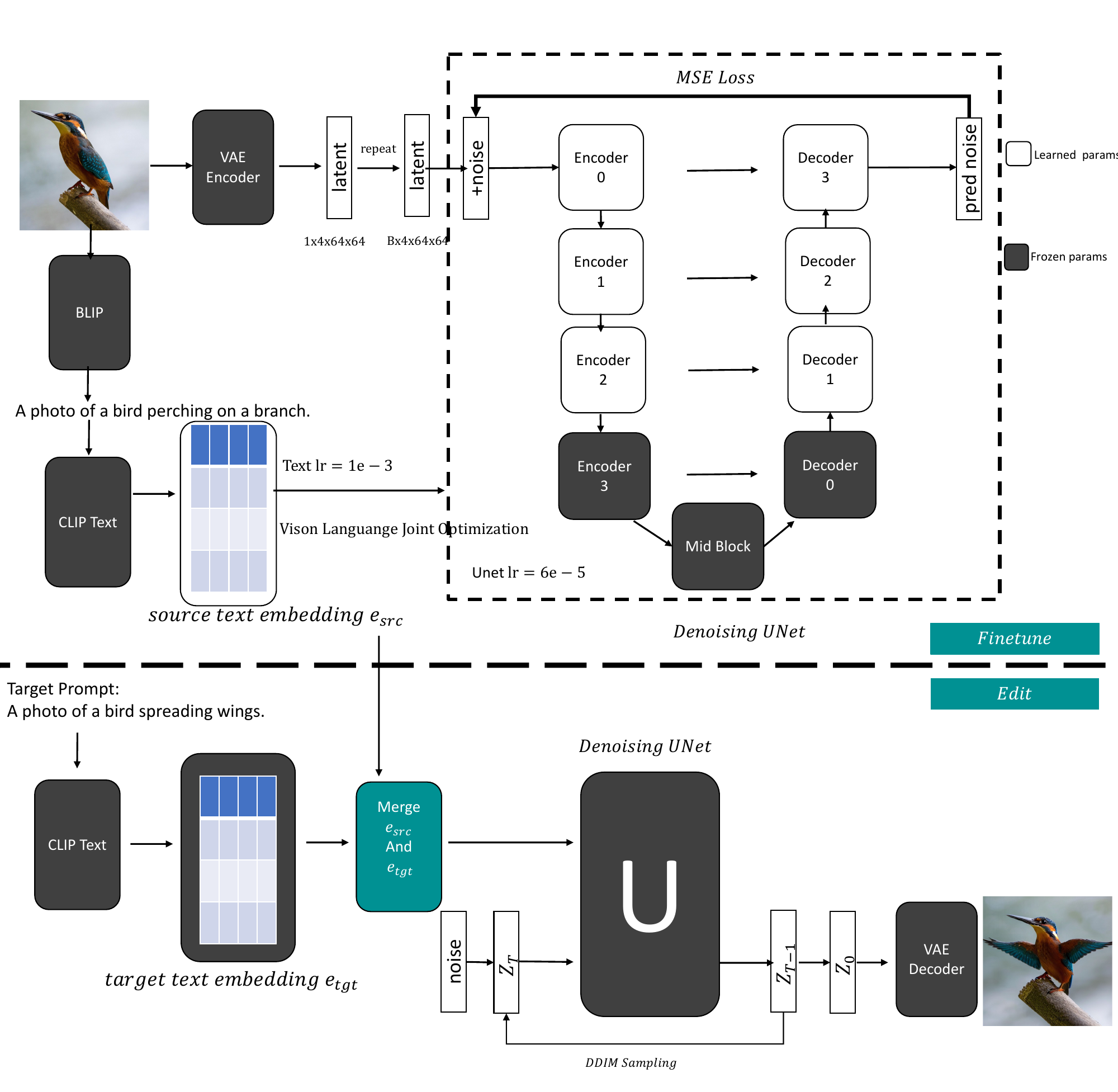}

   \caption{Overall framework of our Forgedit, consisting of a vision-language joint fine-tuning stage and an editing stage. We use BLIP to generate a text description of an original image, and compute an embedding of the source text $e_{src}$ using a CLIP text encoder. The source embedding $e_{src}$ is then jointly optimized with UNet using different learning rates for text embedding and UNet, where the deep layers of UNet are frozen. During the editing process, we merge the source embedding $e_{src}$ and the target embedding $e_{tgt}$ with vector subtraction or projection to get a final text embedding $e$. With our forgetting strategies applied to UNet, we utilize DDIM sampling to get the final edited image.}
   \vspace{-10pt}
   \label{main}
\end{figure*}

\subsection{Joint fine-tuning}
In order to tackle such challenging text-guided image editing problems, we have to fine-tune the model to learn the concepts from the original image thus the model could preserve them consistently during the editing process. It is worth noting that although DDIM inversion \cite{DBLP:conf/iclr/SongME21} could reconstruct the original image, the given text prompt has to be an empty string. If the given text prompt is not empty, DDIM inversion  is incapable to reconstruct original image precisely and often leads to significant appearance shift \cite{DBLP:conf/iclr/HertzMTAPC23,Meng2021SDEditGI}. Thus it is necessary to optimize the network for high quality reconstruction and semantic understanding.
As shown in Figure \ref{main}, we introduce the overall design of our vision-language joint optimization framework.

{\bf Source prompt generation.} We first use BLIP \cite{Li2022BLIPBL} to generate a caption describing the original image, which is referred to as the source prompt. The source prompt is then fed to the text encoder of Stable Diffusion \cite{Rombach2021HighResolutionIS}, generating an embedding $e_{src}$ of source prompt.
Previous three-stage editing method Imagic \cite{DBLP:conf/cvpr/KawarZLTCDMI23} regards target prompt text embedding as source one $e_{src}$.  We found that it is essential to use the BLIP caption instead of using the  target prompt as a pseudo source prompt like Imagic. Otherwise such fine-tuning methods easily lead to overfitting issues, as demonstrated in the 5th column of Figure \ref{compare}.

{\bf Vision-language joint learning.} We choose to optimize encoder layers of 0, 1, 2 and decoder layers of 1, 2, 3 in the UNet structure since we found that fine-tuning deepest features would lead to overfitting in our Forgedit framework.  Similar with Imagic, we regard source text embedding as parameters of the network. Yet different with Imagic, we found it vital to jointly optimize the source text embedding and UNet parameters, which is of great importance for faster learning and better reconstruction quality. In particular, due to a large domain gap between text and image, we use different learning rates  for source text embedding ($10^{-3}$) and  UNet ($6 \times 10^{-5}$), with an Adam Optimizer \cite{DBLP:journals/corr/KingmaB14}. 
For faster training, since we only have a single training image, we repeat the tensors on batch dimension for batch-wise optimization with a batch size of 10.  We use mean square error loss, and empirically found that stable reconstruction results can be achieved when the final loss is less than 0.03. 
With a batch size set to 10, the models are fine-tuned for 35 to 40 steps. 
We stop the training over 35 steps when the loss is less than 0.03, or stop at 40 steps at most. This fine-tuning process is significantly more efficient than Imagic, by taking 30 seconds on a single A100 GPU. The training loss is computed as,
\begin{equation}
L = \mathbb{E}_{z_t,\epsilon_t,t,e_{src}} ||\epsilon_t -\epsilon_{\theta,e_{src}}(z_t,t,e_{src})||_2^2
\label{eq joint}
\end{equation}
where the main difference with the training loss presented in \ref{eq loss} is that $e_{src}$ is considered as parameters to optimize.

 \begin{figure*}[t]
  \centering
  \vspace{-10pt}
   \includegraphics[width=0.8\linewidth]{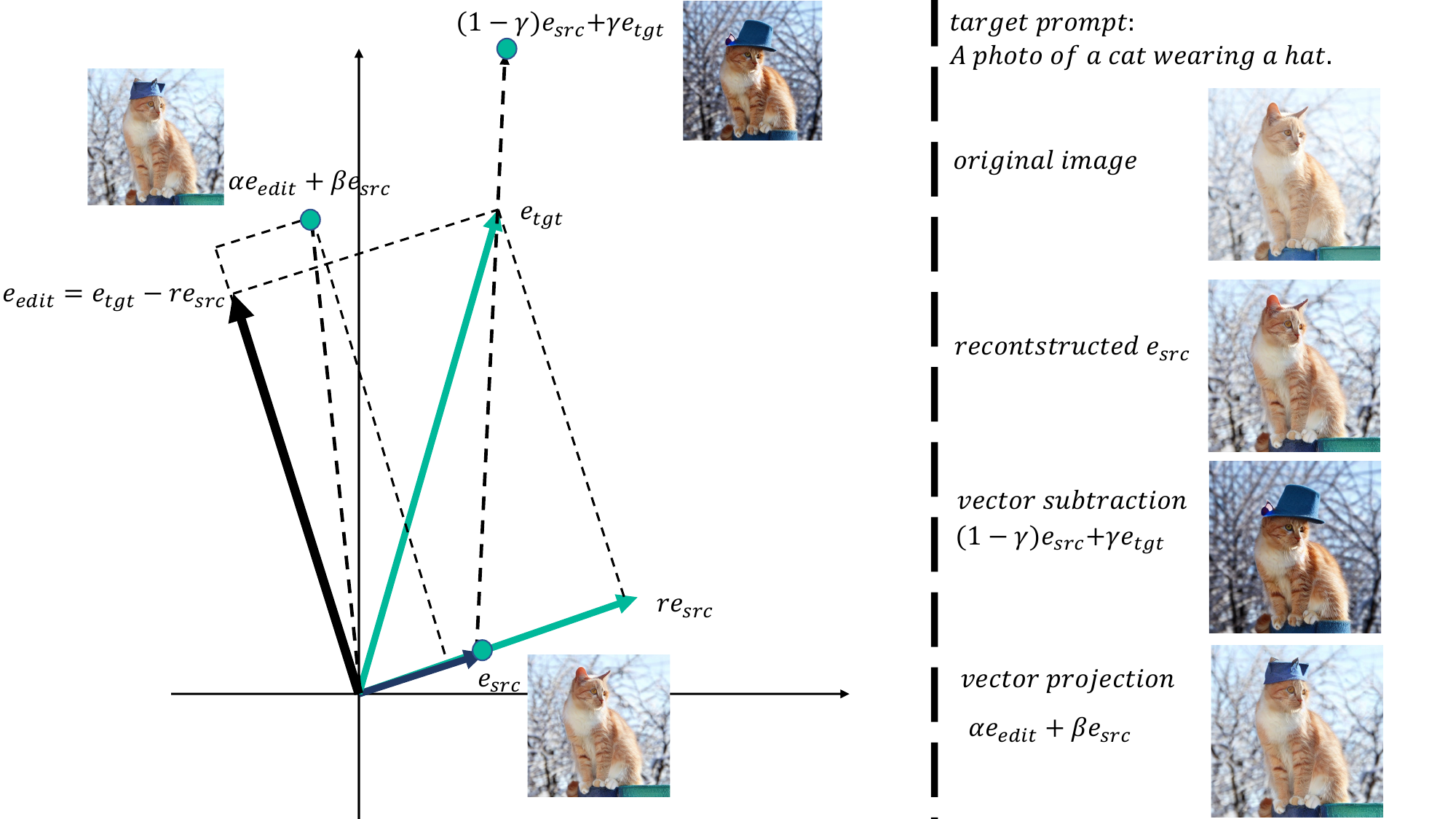}
   \caption{ We demonstrate vector subtraction and vector projection to merge $e_{src}$ and $e_{tgt}$. Vector subtraction could lead to inconsistent appearance of the object being edited since it cannot directly control the importance of $e_ {src}$. The vector projection decomposes the $e_{tgt}$ into $re_{src}$ along $e_{src}$ and $e_{edit}$ orthogonal to $e_{src}$. We can directly control the scales of $e_{src}$ and $e_{edit}$ by summation. }
   \vspace{-20pt}
   \label{reason}
\end{figure*}

\subsection{Reasoning and Editing} 
We first input the target prompt to the CLIP \cite{Radford2021LearningTV} text encoder of the Stable Diffusion model \cite{Rombach2021HighResolutionIS}, computing a target text embedding $e_{tgt}$. With our learned source text embedding $e_{src}$, we propose two methods to compute a combination of $e_{src}$ and $e_{tgt}$ so that the combined text embedding can be applied for editing the original image according to the target prompt. Given $e_{src} \in \mathbb{R}^{B \times N \times C}$ and  $e_{tgt} \in \mathbb{R}^{B \times N \times C}$ , we conduct all vector operations on the $C$ dimension to get the final text embedding $e$.

{\bf Vector Subtraction.}
 We use the same interpolation method as Imagic \cite{DBLP:conf/cvpr/KawarZLTCDMI23}, 
\begin{equation}
e=\gamma e_{tgt}+(1-\gamma) e_{src}=e_{src}+\gamma (e_{tgt}-e_{src})
\label{eq one channel normalized}
\end{equation}

As shown in Figure \ref{reason}, the final text embedding $e$ is obtained by travelling along vector subtraction $e_{tgt}-e_{src}$ . In our experiments, we found that in most cases, $\gamma$ goes beyond 1 when the editing is performed successfully. This leads to a problem that the distance between the final embedding $e$ and the source embedding $e_{src}$ may be so far that the appearance of the edited object could change vastly.

{\bf Vector Projection.}
We propose to use vector projection to better preserve the appearance of the original image. As shown in the Figure \ref{reason}, we decompose a target prompt text embedding $e_{tgt}$ into a vector along $e_{src}$ and a vector orthogonal to $e_{src}$. We call the orthogonal vector  $e_{edit}$. We first calculate the ratio $r$ of the projected vector on $e_{src}$ direction.

\begin{equation}
r=\frac{e_{src}e_{tgt}}{||e_{src}||^{2}}
\label{eq one channel normalized}
\end{equation}

Thus, we could get the $e_{edit}$ by computing
\begin{equation}
e_{edit}=e_{tgt}-re_{src}
\label{eq one channel normalized}
\end{equation}

To preserve more details of original image, we sum $e_{src}$ and $e_{edit}$ with coefficient $\alpha$ and $\beta$,
\begin{equation}
e=\alpha  e_{src}+\beta e_{edit}
\label{eq one channel normalized}
\end{equation}


{\bf Editing.} We use DDIM sampling \cite{DBLP:conf/iclr/SongME21} with a classifier free guidance \cite{Ho2022ClassifierFreeDG} to conduct the edit. The guidance scale is 7.5. For vector subtraction, we iterate over a range of $\gamma \in [0.8,1.6]$. For vector projection, we choose $\alpha$ from two values $\{0.8,1.1\}$, and $\beta$ from a range of [1.0,1.5]. 
 \begin{figure*}[t]
  \centering
  \vspace{-10pt}
   \includegraphics[width=0.8\linewidth]{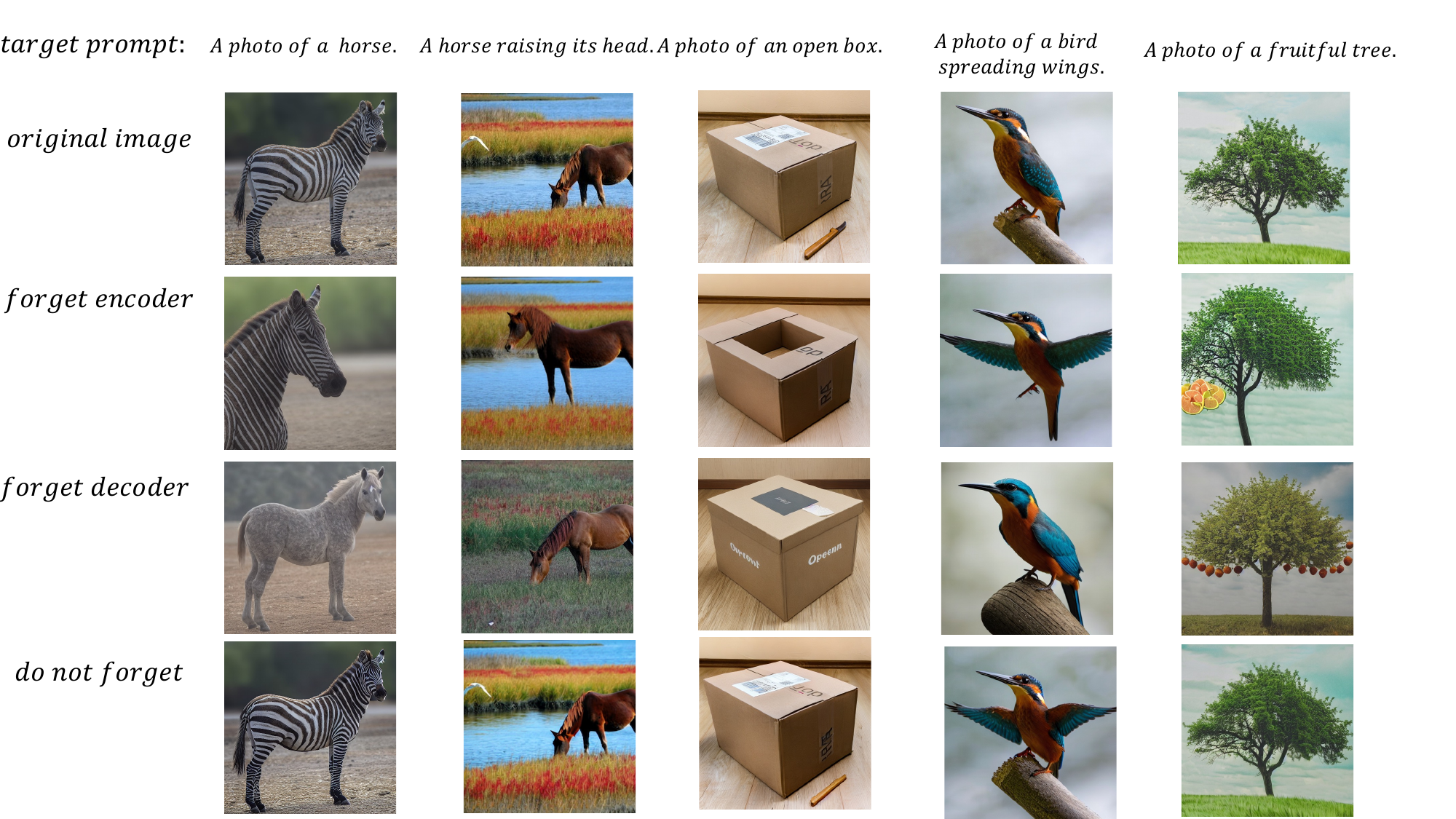}
   \caption{The encoder of UNets learn features related to pose, angle, structure and position. The decoder are related to appearance and texture. Thus we design a forgetting strategy according to the editing target. 
   }
   \label{forget}
\end{figure*}
\subsection{Forgetting Strategy}
{\bf Forgetting mechanism} In some cases the network still overfits since there is only one training image provided. The fine-tuning process is computational expensive compared to sampling process, thus we design a forgetting strategy during sampling process to tackle the overfitting problem. The network is only fine-tuned once, and can be converted to multiple different networks during sampling process by merging certain fine-tuned parameters $w_{learned}$ and the corresponding parameters of original UNet (before fine-tuning) $w_{orig}$, with a balance coefficient $\sigma$. 
In practice, we found that $\sigma=0$ works in general, which means that we can simply replace the fine-tuned parameters with original parameters so that the network completely forgets these learned parameters.
\begin{equation}
w=\sigma w_{learned}+(1-\sigma) w_{orig}
\label{eq one channel normalized}
\end{equation}

{\bf UNet's property}  As shown in Figure \ref{forget}, we found an interesting property of UNets in diffusion models. The encoder of UNets learns space and structure information like the pose, action, position, angle and overall layout of the image, while the decoder learns appearance and identity instead. 

{\bf Forgetting strategy} If the target prompt tends to edit space information, for example, the pose or layout, we choose to forget parameters of the encoder. If the target prompt aims to edit the appearance, the parameters of decoder should be forgotten. Currently we only apply the forgetting strategy when a text embedding $e$ is obtained by vector subtraction in previous section.  For editing with the forgetting strategy, we iterate over a range of $\gamma \in [0.0,1.4]$. For different settings of forgetting strategy, we explore their effects in ablation study, as shown in Figure \ref{forgetdecoder} and Figure \ref{forgetencoder}.
\begin{figure*}[t]
  \centering
  \vspace{-10pt}
   \includegraphics[width=0.8\linewidth]{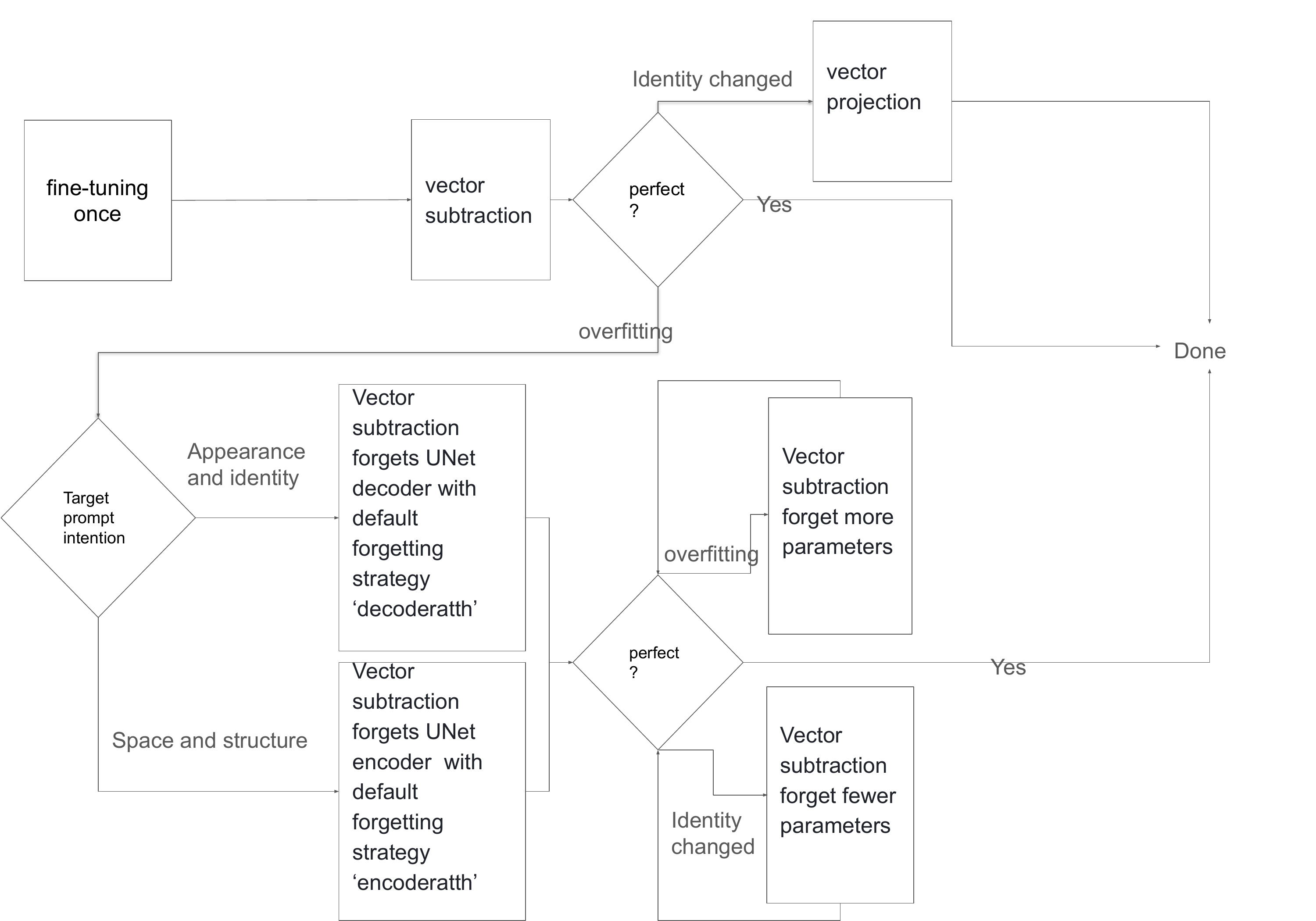}
   \caption{Forgedit Workflow. 
   }
   \vspace{-10pt}
   \label{flow}
\end{figure*}
\subsection{WorkFlow and Limitations}
The overall workflow of Forgedit is explained in Figure \ref{flow}. The fine-tuning stage is the same for all images. The diamonds in the figure indicate that the process depends on the users's choices and preferences. In practice, these user decisions can also be replaced by thresholds on  metrics like CLIP score and LPIPS score, for completely automatic editing.

\begin{figure*}[h!]
  \centering
    
   \includegraphics[width=0.8\linewidth]{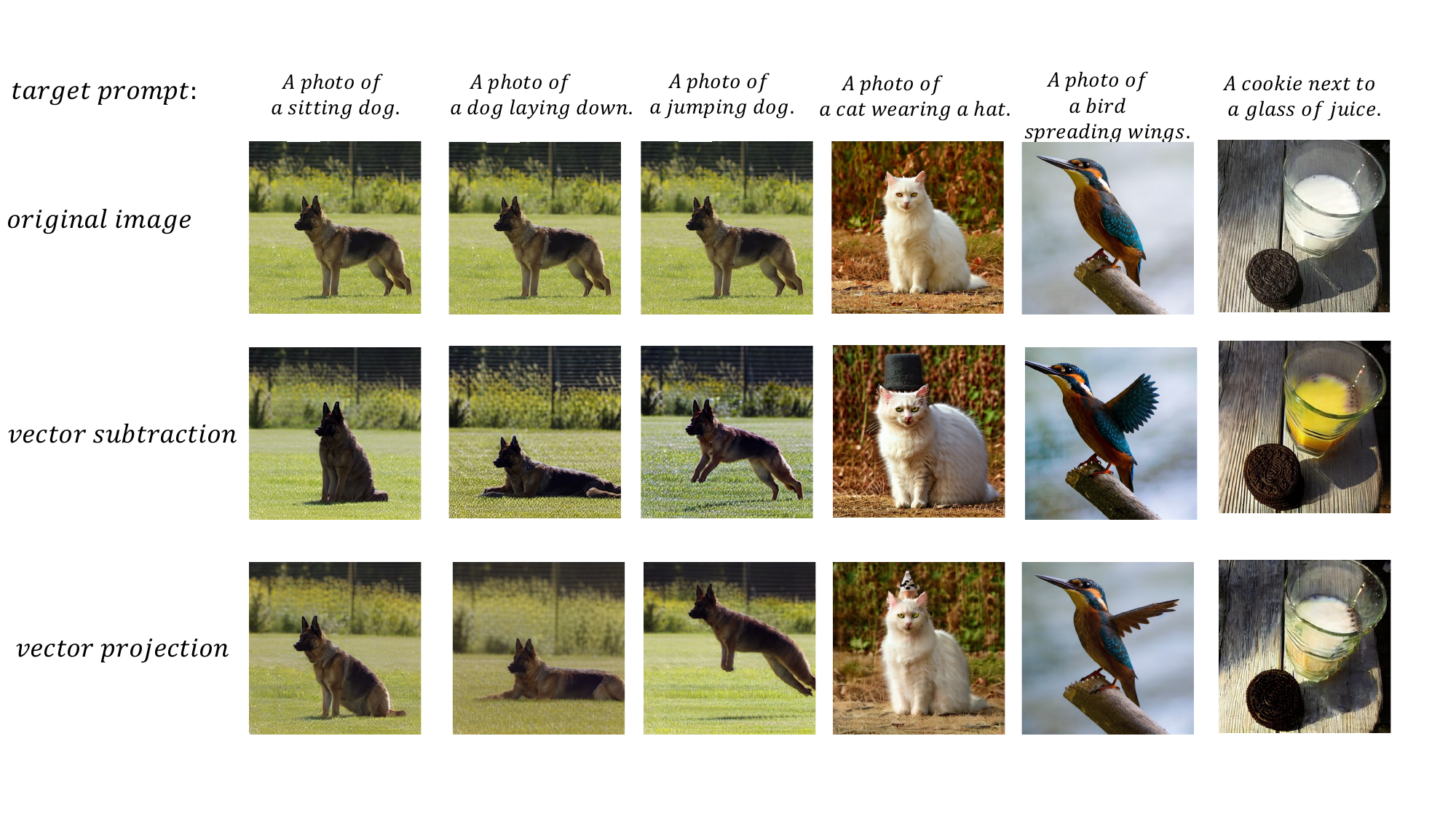}
  \vspace{-10pt}
   \caption{Comparisons of vector subtraction and vector projection, which are  complementary. 
   }
   \vspace{-10pt}
   \label{reasonablation}
\end{figure*}
\section{Experiments}
{\bf Benchmark.} TEdBench \cite{DBLP:conf/cvpr/KawarZLTCDMI23} is one of the most difficult public-available text-guided image editing benchmarks. It contains 100 editings, with one target prompt and one image for each edit. These target prompts are very general with diversity, including but not limited to changing the appearance of objects, replacing certain parts of the image, changing the position, action and number of the object, editing multiple objects with complex interactions. In particular, non-rigid edits turn out to be very tough for many SOTA text-guided image editing methods. In terms of quantitative evaluation, we utilize CLIP Score \cite{DBLP:conf/emnlp/HesselHFBC21} to measure semantic alignments with target prompt, and LPIPS score \cite{Zhang2018TheUE} and FID score \cite{DBLP:conf/nips/HeuselRUNH17} to indicate fidelity to the original image.

\subsection{Ablation Study}



{\bf Vector subtraction vs vector projection.}
We compare two different reasoning methods to merge $e_{src}$ and $e_{tgt}$ to get the final text embedding $e$, shown in Figure \ref{reasonablation} . 
%
These two methods are complementary to each other, with vector projection better at preserving the identity, and vector subtraction better at replacing objects.

\begin{figure*}[t]
  \centering
  \vspace{-10pt}
   \includegraphics[width=0.8\linewidth]{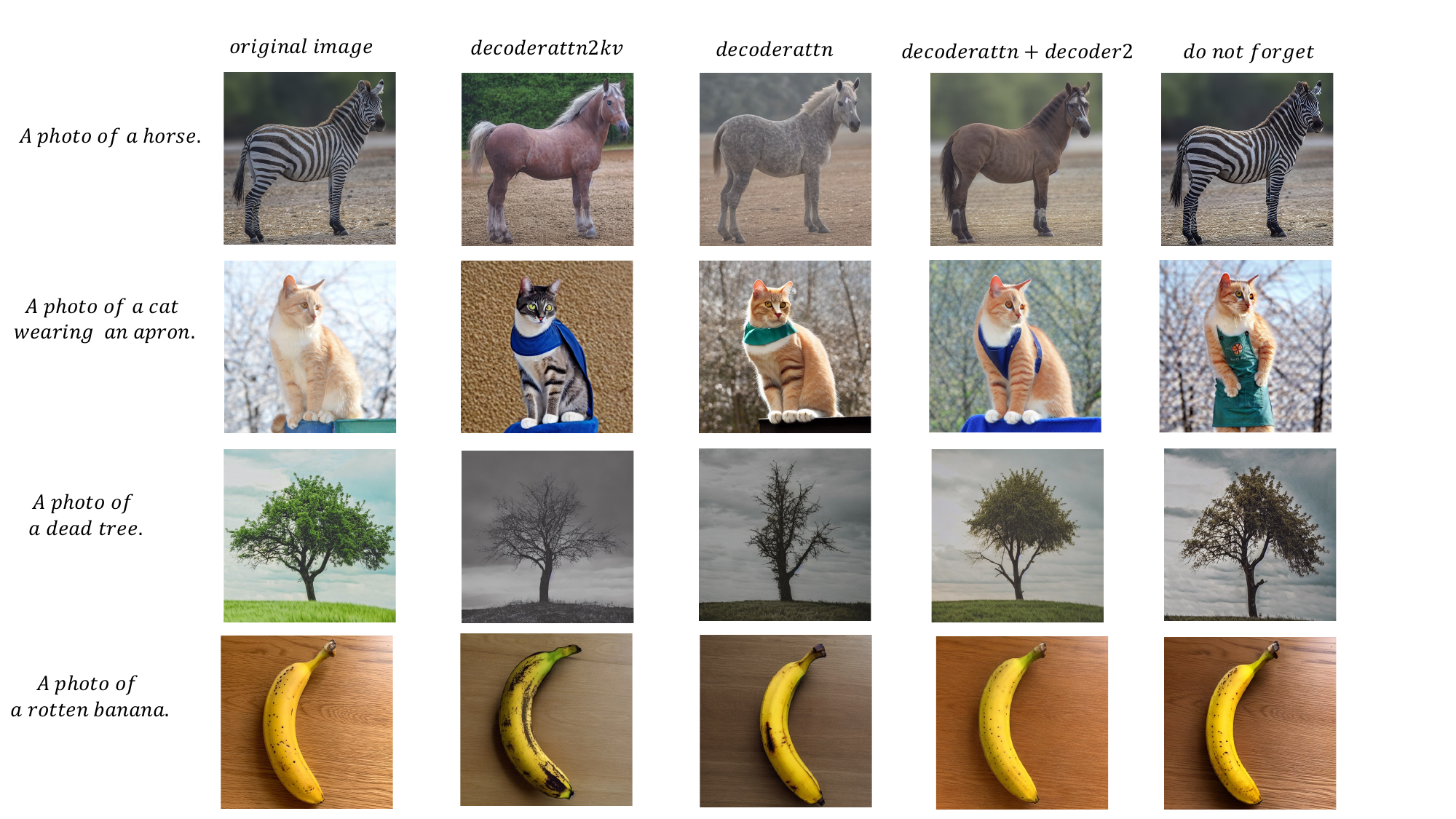}

   \caption{ We explore various forgetting strategies for decoder. All learned encoder parameters are preserved. In the $2^{nd}$ to $4^{th}$ columns, we preserve decoder cross-attention parameters, decoder self-attention and cross-attention, decoder self-attention, cross-attention and the entire decoder2 block, forgetting all the other parameters of decoder.
   }
   \label{forgetdecoder}
\end{figure*}

\begin{figure*}[t]
  \centering
  \vspace{-10pt}
   \includegraphics[width=0.8\linewidth]{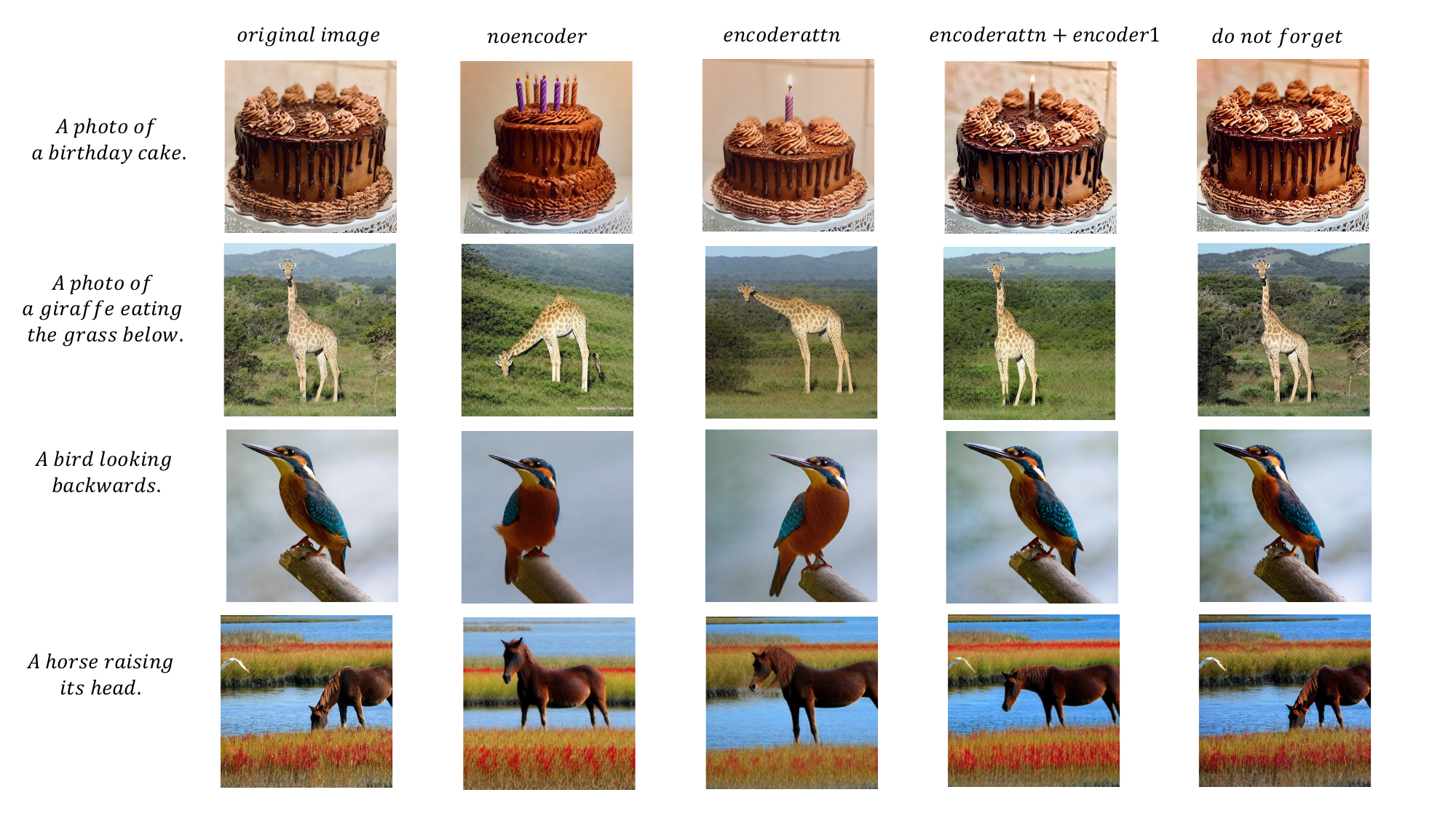}

   \caption{ We explore different forgetting strategies for encoder. All learned decoder parameters are preserved. For the second to fourth column each, we preserve none of the encoder parameters, encoder self attention and cross attention, encoder self attention and cross attention and the entire encoder1 block, forgetting all the other parameters of encoder.
   }
   \label{forgetencoder}
\end{figure*}

{\bf Forgetting strategy.} We first inference without forgetting strategies. If overfitting happens, we choose from the default 'encoderattn' or 'decoderattn' strategy according to the UNet property and target prompt intention. The 'encoderattn' means forgetting all encoder parameters except attention-related parameters. 'decoderattn' means forgetting all decoder parameters except attention-related parameters. The user may choose to forget more or fewer parameters according to the editing results, which we demonstrate and explain in \ref{forgetencoder} and \ref{forgetdecoder}.

\begin{figure*}[h!]
  \centering
   \includegraphics[width=0.5\linewidth]{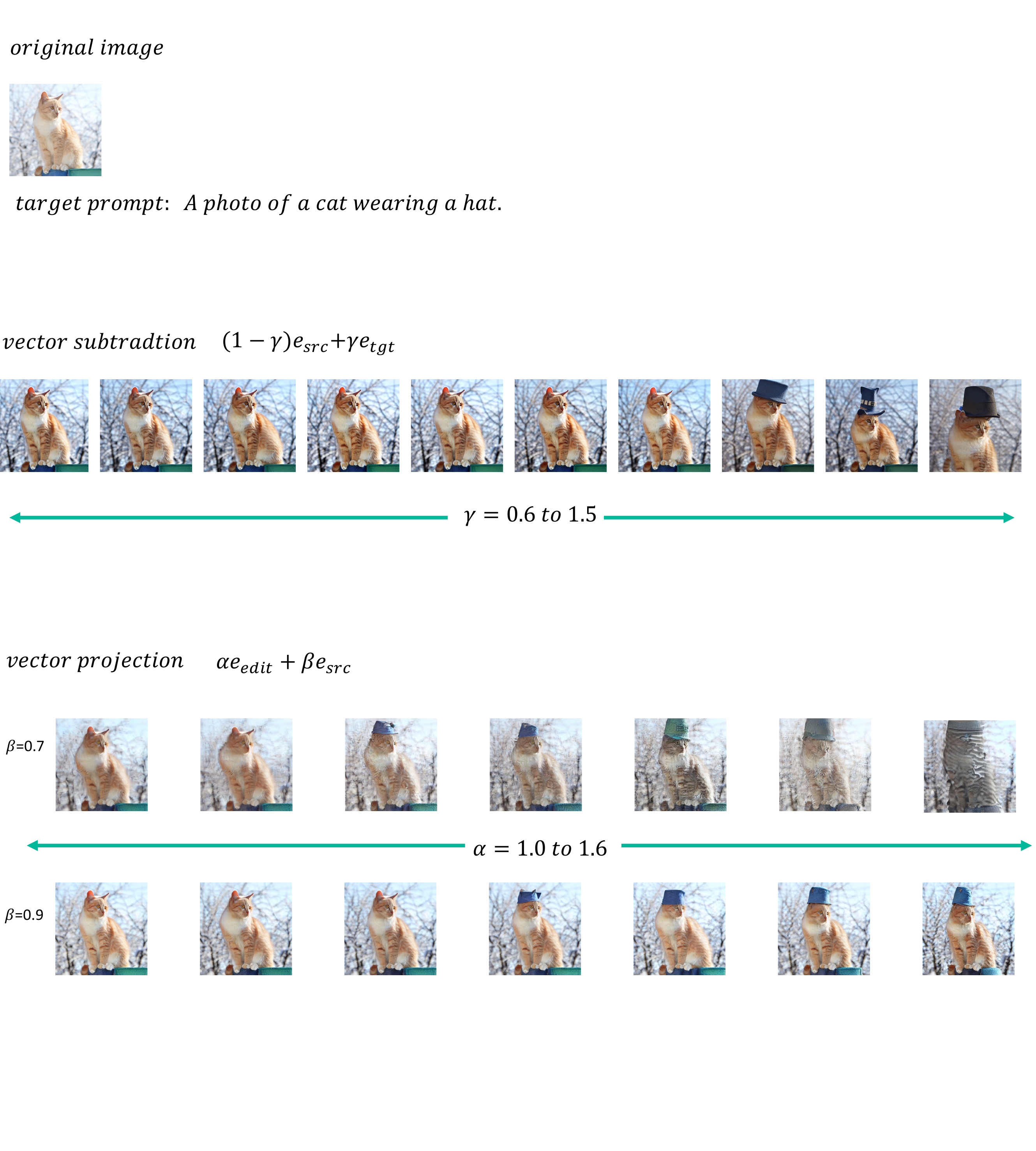}
\vspace{-20pt}
   \caption{ $\gamma$ for vector subtraction and $\alpha$, $\beta$ for vector projection.
   }
   
   \label{interpolation}
\end{figure*}

{\bf hyperparameters effects for text embedding interpolation}
We explore the effects of hyperparameters in vector subtraction and vector projection in Figure \ref{interpolation}.

\begin{figure*}[h!]
 
  \centering
  
   \includegraphics[width=0.5\linewidth]{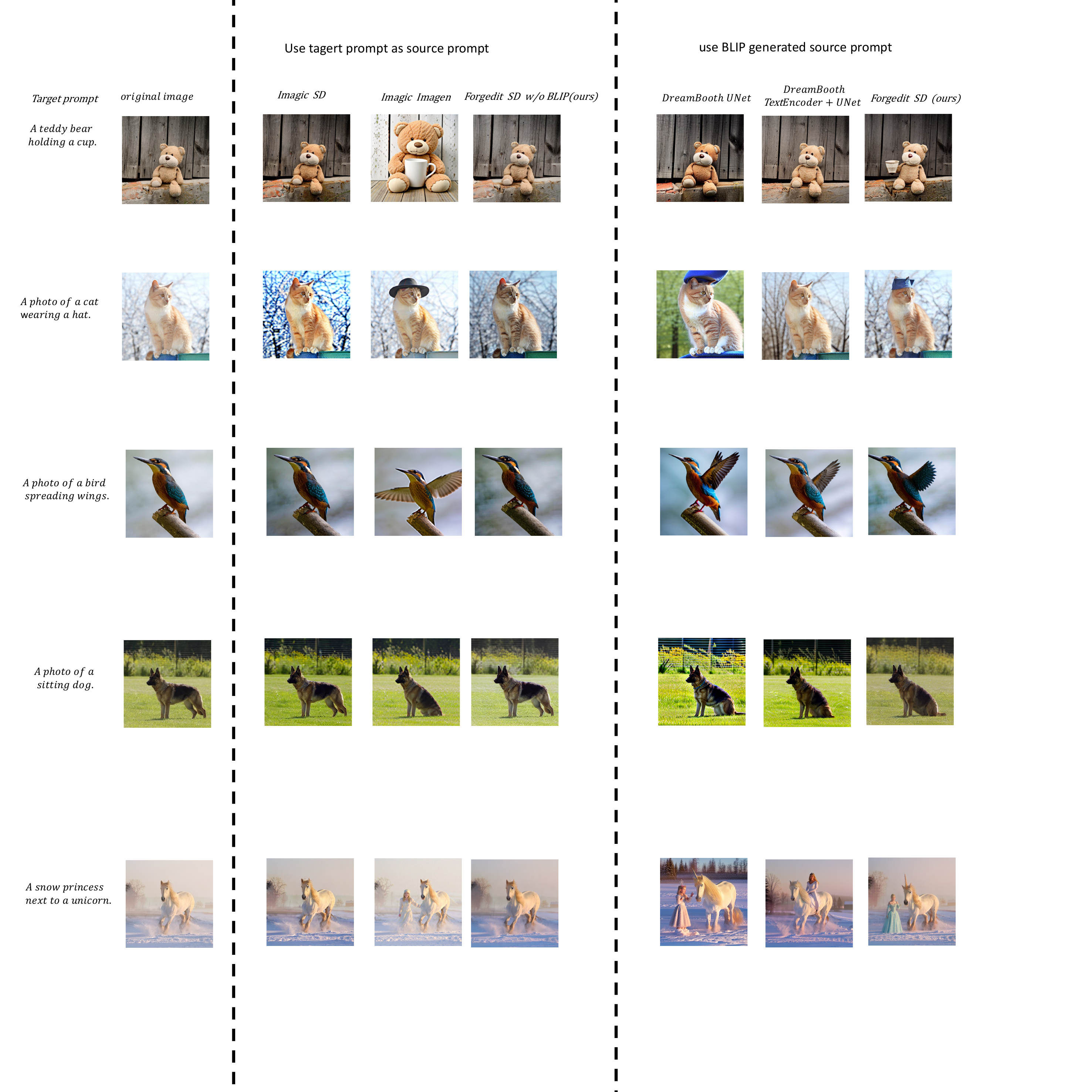}

   \caption{ What should the source prompt be? Excluding the usage of forgetting strategies for ablation, we could find that Forgedit using target prompt leads to severe overfitting, yet Forgedit using BLIP generated source prompt eases overfitting.
   }
   
   \label{blip}
\end{figure*}

{\bf What should source prompt be?}
We explore the importance of using generated caption as source prompt. We use BLIP generated source prompt to describe the original image,  yet previous SOTA method Imagic uses target prompt  as source prompt. Since target prompt indicates the editing target, it is obviously inconsistent with the original image. Although target prompt embedding is optimized with the original image for reconstruction, it is carefully controlled by learning rate and made sure not far from the initial target prompt embedding. Thus in many cases, using target prompt as source prompt confuses the editing model to correctly understand the editing intention of target prompt, making it lose the ability to conduct the edit,  even after finetuning.  In Figure \ref{blip}, we show cases that do not need to use forgetting strategy from Figure \ref{compare}, so that we could remove the effects of forgetting strategy. If target prompt is used instead of BLIP generated source prompt, all these cases of Forgedit without using generated source prompt  will overfit.


\subsection{Comparison with State-of-the-art}
\begin{figure*}[h!]
  \centering
  \vspace{-10pt}
   \includegraphics[width=1\linewidth]{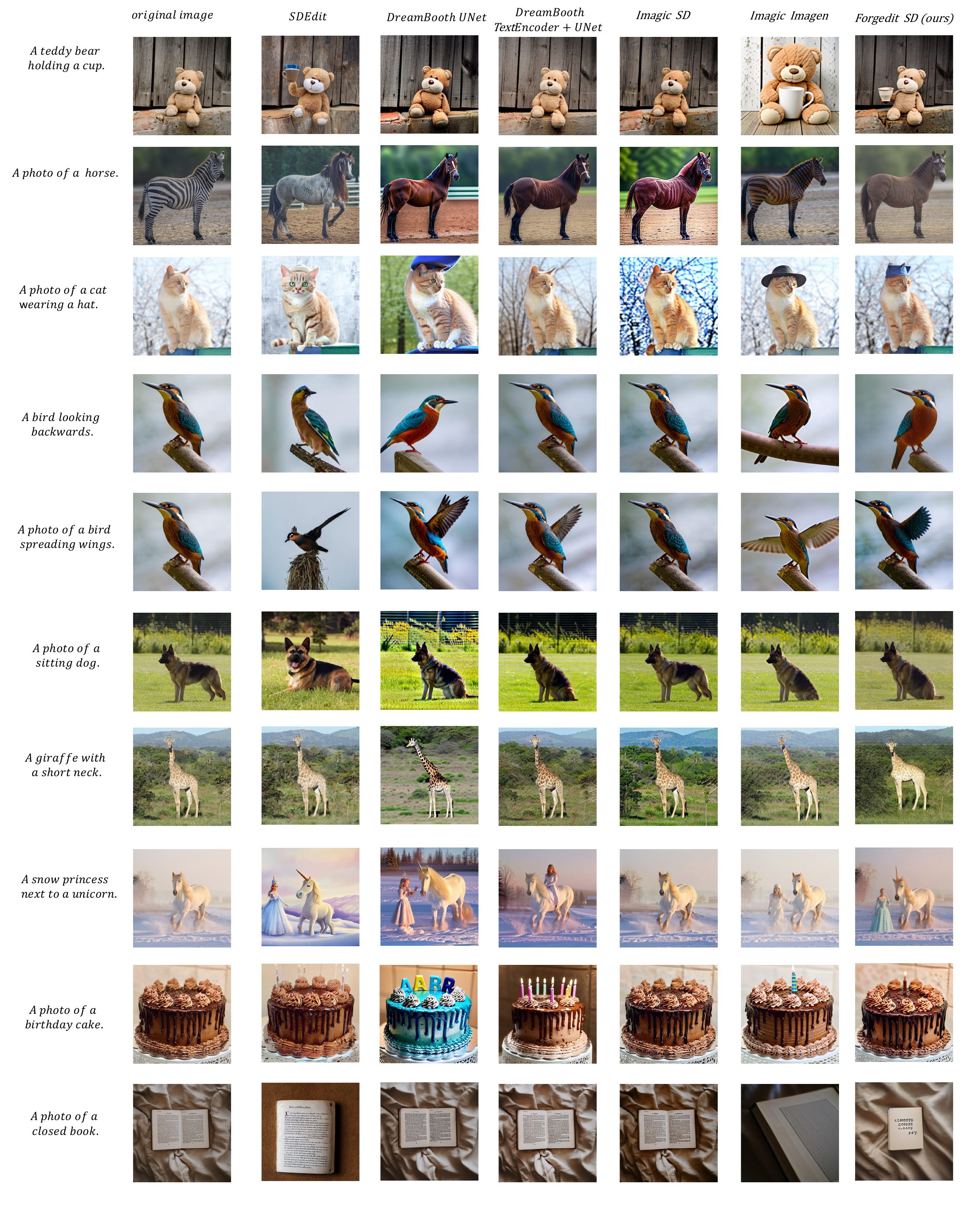}
 
   \caption{Comparison with SOTA: non-optimization SDEdit, optimization BLIP+DreamBooth and Imagic, demonstrating the strong editing ability and stable identity preservation of Forgedit.
   }
   \label{compare}
\end{figure*}

We compare qualitative editing results of our Forgedit with SOTA methods on several random test samples from TEdBench in Figure \ref{compare}, demonstrating stronger semantic alignments with target prompts and more precise identity preservation than other methods. Quantitatively, our Forgedit with the even outdated Stable Diffusion 1.4,  surpasses the current SOTA Imagic+Imagen  on TEdBench benchmark in terms of both CLIP Score, LPIPS Score and FID Score, shown in Table 
\ref{tablecompare}. For FID score, we follow the advice of the authors by setting dimension to 192 since there are only 100 samples in TEdBench.

\begin{table}[t]

\begin{center}
\begin{tabular}{c|c|c|c}
{Editing method} & {CLIP Score $\uparrow$} & {LPIPS Score $\downarrow$} & {FID Score $\downarrow$}\\
 \hline 
Imagic+Imagen \cite{DBLP:conf/cvpr/KawarZLTCDMI23} & 0.748 & 0.537 & 8.353\\
Forgedit+SD (ours) & {\bf 0.771} & {\bf 0.534} &{\bf 7.071}\\


\end{tabular}
\end{center}
\caption{Our Forgedit with Stable Diffusion is the new state-of-the-art text-guided image editing method on the challenging benchmark TEdBench, surpassing previous SOTA Imagic+Imagen. }
\vspace{-20pt}
\label{tablecompare}
\end{table}

\subsection{Visual Storytelling}
Our Forgedit could precisely preserve the characteristics of actors and is capable of conducting complex non-rigid editing, which makes our Forgedit an ideal tool for visual storytelling and long video generation with strong consistency and very arbitrary scene and action. In Figure \ref{movie}, we input a random image generated by SDXL \cite{DBLP:conf/nips/HeuselRUNH17} and then use Forgedit with Realistic Vision V6.0 B1 noVAE, a variant of Stable Diffusion to generate various samples for different target prompts. With image to video models, for example Stable Video Diffusion \cite{blattmann2023stable}, we could generate movies with high consistency of several minutes.

\section{Conclusion}

We present our Forgedit framework to tackle the challenging text-guided image editing problem. Forgedit solves the overfitting problem of Diffusion Models when fine-tuning with only one image, and obtain new SOTA on TEdBench. Forgedit could be used for precise text-guided image editing and flexible controllable image generation. 

\section{appendix}

\subsection{BLIP+DreamBooth}
 In order for DreamBooth \cite{DBLP:conf/cvpr/RuizLJPRA23}  to be applied in text guided image editing, we utilize BLIP \cite{Li2022BLIPBL} to generate captions describing the original image like our Forgedit. With the caption, we train the UNet to reconstruct the original image and edit the image by directly using the target prompt to guide the fine-tuned UNet for image generation, shown in the 3rd column of Figure \ref{dreambooth}. We also experiment with an improved version by training text encoder and UNet at the same time, shown in the 4th column. Such simple fine-tuning of UNet and text encoder are actually very powerful text guided image editing methods. Our BLIP+DreamBooth uses BLIP generated caption, different with the fact that original DreamBooth requires a user provided caption in a special form of 'a [V] object' referring the object to be reconstruct. Following the settings of DreamBooth \cite{DBLP:conf/cvpr/RuizLJPRA23}, we use a learning rate of $5\times 10^{-6}$ for both text encoder and UNet, with a batch size of 4. We train BLIP+DreamBooth with one image for 100 steps, which takes more than one minute on a A100 GPU.  Unlike original DreamBooth which needs 3 to 4 images to learn the new object concept, we found that with BLIP+DreamBooth one image is enough to reconstruct the majority features of the original image. However, BLIP+DreamBooth, when only UNet is fine-tuned, suffers from underfitting since it cannot preserve the identity of the objects in many cases. BLIP+DreamBooth suffers from overfitting in many cases when text encoder and UNet are jointly fine-tuned. In fact, we found that our Forgedit can also be simply adapted to help tackling such overfitting issues of BLIP+DreamBooth, shown in the \ref{dreamboothforgedit}, which again demonstrates the strong generalization of Forgedit framework on various optimization based editing methods.
 \label{blip+dreambooth}
\subsection{ DreamBooth+Forgedit}
Forgedit is a very general framework, whose main features come from three aspects: joint vision and language learning with original image, obtaining final text embedding by vector subtraction and vector projection, using forgetting strategies to tackle the overfitting issues. Here we show how to extend Forgedit to BLIP+DreamBooth \cite{Li2022BLIPBL,DBLP:conf/cvpr/RuizLJPRA23}. It is also possible to adapt our Forgedit to other Diffusion Models \cite{Ho2021CascadedDM,DBLP:conf/nips/SahariaCSLWDGLA22} or fine-tuning methods \cite{DBLP:conf/iclr/HuSWALWWC22}, which we will explore in the future.\\
\textbf{vision and language joint learning } This is natural for the method which we call BLIP+DreamBooth Text Encoder and UNet, since the Text Encoder and UNet are jointly trained already.\\
\textbf{vector subtraction and vector projection} Our Forgedit presented in the main paper regards the source text embedding as a part of the network to optimize. For BLIP+DreamBooth, since we have already fine-tuned the text encoder, we switch to use the text encoder to get source text embedding directly from source prompt. Now we can use vector subtraction and vector projection in the same way.\\
\textbf{forgetting strategy} We could directly apply the forgetting strategies to BLIP +DreamBooth. However, since the information are injected into both text encoder and UNet, our forgetting strategies on UNet may still fail in some cases. We will explore the forgetting strategies in text encoder in the future.

\begin{figure*}[h!]
  \centering
   \includegraphics[width=0.8\linewidth]{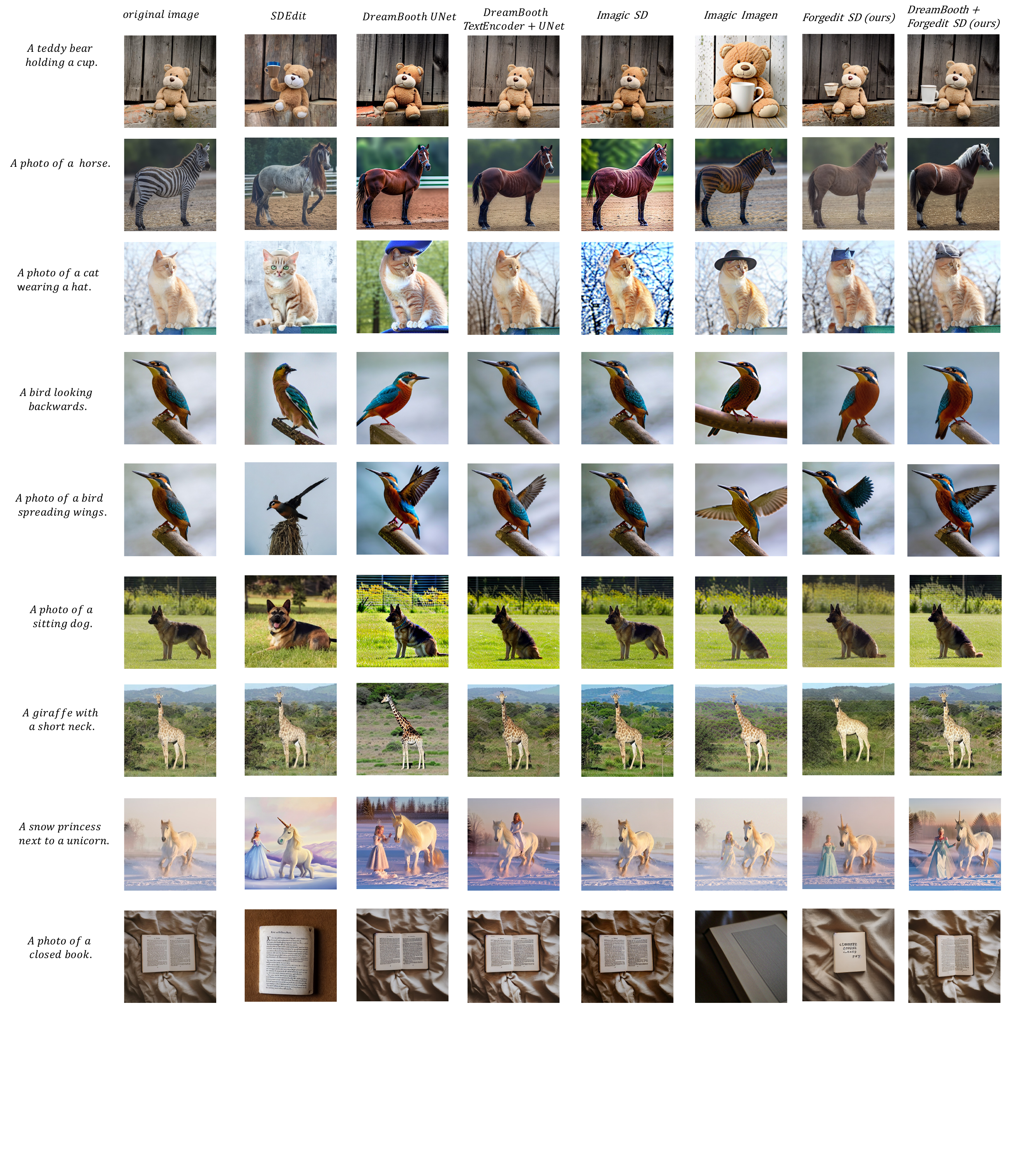}
    \vspace{-20pt}
   \caption{ The editing effects of DreamBooth+Forgedit and comparisons with SOTA methods. Please note that DreamBooth+Forgedit is not always better than Forgedit, as shown in the last two columns.
   }
   \label{dreambooth}
\end{figure*}

We show some cases in Figure \ref{dreambooth}, and compare the editing effects of DreamBooth+Forgedit with previous state-of-the-art text guided image editing methods and our vanilla Forgedit presented in the main paper. Comparing the 4th column and the last column, we could find that with Forgedit framework, the editing ability of DreamBooth has been tremendously improved and the overfitting issues are solved in most cases. Please note that  DreamBooth+Forgedit is not a simple combination of our vanilla Forgedit presented in the main paper and BLIP+DreamBooth, since the fine-tuning process of our vanilla Forgedit is different with DreamBooth+Forgedit. This leads to the fact that DreamBooth+Forgedit is not always better than vanilla Forgedit, which are shown in the last two columns in Figure \ref{dreambooth}. 

\label{dreamboothforgedit}
\begin{figure*}[h!]
  \centering
   \includegraphics[width=0.5\linewidth]{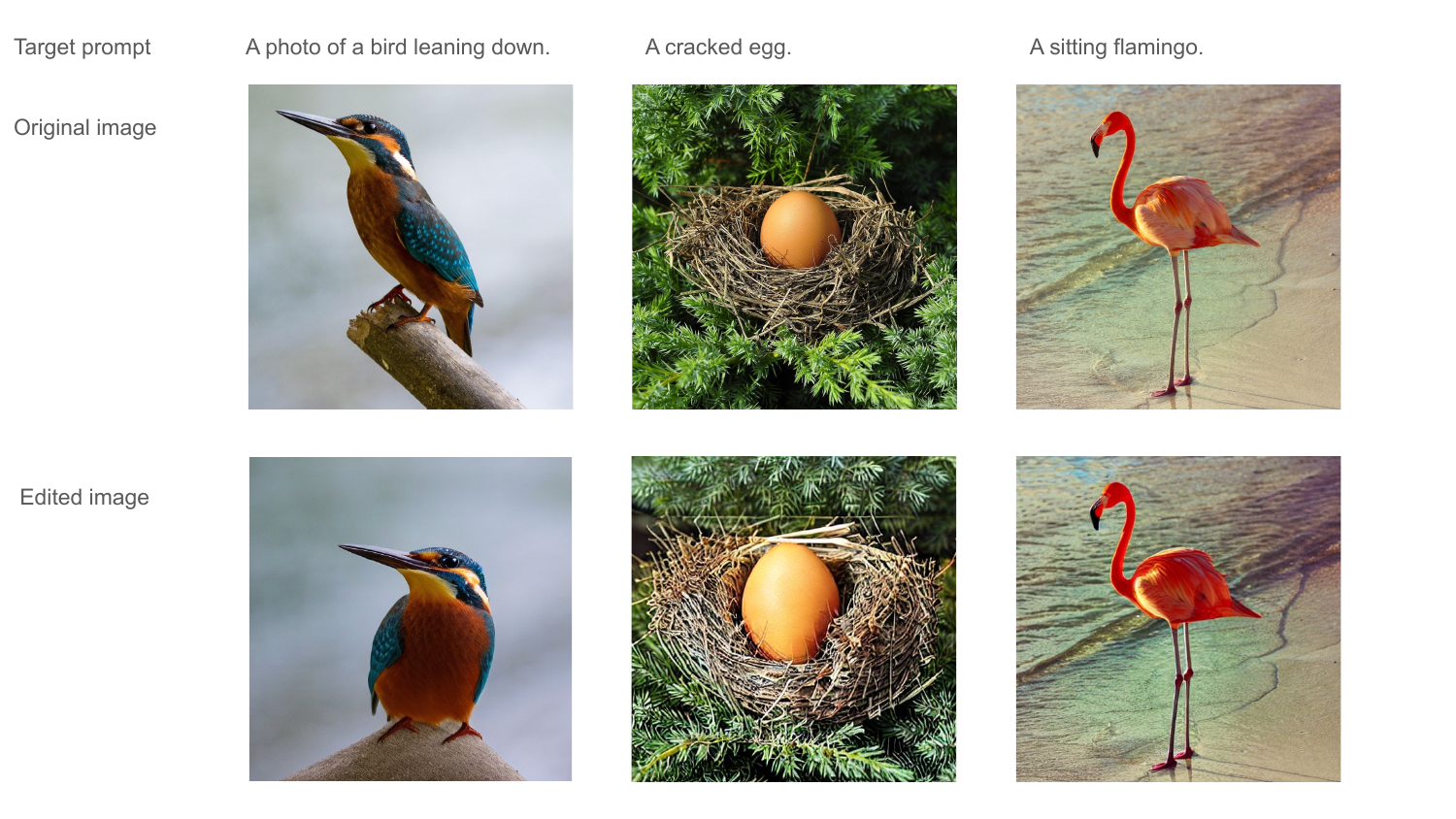}

   \caption{ Bad cases from TEdBench.
   }
   \label{badcase}
\end{figure*}

\subsection{Limitations}

First  the effect of Forgedit is influenced by randomness. The fine-tuning process inevitably introduces randomness thus for some particular cases, we cannot guarantee to perfectly reconstruct the details of original image thus we have to run the fine-tuning stage several times for these challenging cases.  The sampling procedure is also related to the initial random seed of reverse process, thus for some  extremely challenging cases we have to sample tens of images or even hundreds, though rarely the case, before we could get a proper edited one.

Second, the editing capability of Forgedit is restricted by the utilized Diffusion Model. If the target prompt cannot even be generated by the Diffusion Model itself, it is almost impossible to accomplish the edit according to the target prompt. For example, the prompt 'a sitting flamingo' cannot be generated by Stable Diffusion at all, thus Forgedit cannot successfully edit it either. Such an issue could possibly be solved by switching to better Diffusion Models.

We show some typical bad cases in Figure \ref{badcase}.
\label{limitation}

\clearpage  

%
%
\bibliographystyle{splncs04}
\bibliography{eccvbib}
\end{document}